\documentclass{article} 
\usepackage{iclr2027_conference,times}

\usepackage{amsmath,amsfonts,bm}

\def\eqref#1{equation~\ref{#1}}

\def\1{\bm{1}}

\DeclareMathAlphabet{\mathsfit}{\encodingdefault}{\sfdefault}{m}{sl}
\SetMathAlphabet{\mathsfit}{bold}{\encodingdefault}{\sfdefault}{bx}{n}

\usepackage{hyperref}
\usepackage{url}
\usepackage{booktabs} 
\usepackage{xcolor}
\usepackage{colortbl}
\usepackage{tabularx}
\definecolor{TableBlue}{HTML}{E7EFF8}
\definecolor{TablePeach}{HTML}{FBF0E5}
\definecolor{TableGray}{HTML}{E9EBF0}
\usepackage{float}

\usepackage{graphicx}
\usepackage{wrapfig}
\usepackage{needspace}

\title{Mind the Execution Gap: Action-Semantic Mismatch in World-Model Control}

\author{
Shengtao Wen\textsuperscript{$\heartsuit$}, Xiang Chen\textsuperscript{$\heartsuit$*}, Yu Tian\textsuperscript{$\spadesuit$}, Lingbing Guo\textsuperscript{$\diamondsuit$}, 
Lina Gong\textsuperscript{$\heartsuit$}, Sheng-Jun Huang\textsuperscript{$\heartsuit$} \\[1ex]
\textsuperscript{$\heartsuit$}Nanjing University of Aeronautics and Astronautics, Nanjing, China \\
\textsuperscript{$\spadesuit$}Tsinghua University, Beijing, China \\
\textsuperscript{$\diamondsuit$}Nanjing University, Nanjing, China \\[1ex]
\texttt{\{shengtao\_wen, xiang\_chen, gonglina\}@nuaa.edu.cn,} \\
\texttt{tianyu181@mails.ucas.ac.cn} \\
\texttt{lbguo@nju.edu.cn}
}

\iclrfinalcopy 
\begin{document}
\raggedbottom

\maketitle


\begin{abstract}
World-model controllers rely on action-conditioned dynamics for prediction and planning, yet real control systems often execute commands asynchronously due to communication delay, packet loss, reordering, and actuator buffering.
We study how asynchronous execution changes the action semantics assumed within world-model controllers, rather than treating it only as an external control disturbance.
Through controlled interventions, we identify two architecture-dependent failure modes: planning-based controllers such as TD-MPC2 suffer from a future-action timeline mismatch between imagined and executed action sequences, while recurrent world models such as DreamerV3 can attribute observed transitions to commands that were not actually applied.
Our analysis shows that TD-MPC2 requires the correct future action sequence during latent dynamics rollout, whereas DreamerV3 requires timely attribution of each transition to the action that generated it.
Based on these findings, we introduce two lightweight execution-consistent interfaces, Future-Sequence for TD-MPC2 and Applied-Action Feedback for DreamerV3, that correct these mismatches without modifying the pretrained world models.
Experiments across delays, packet loss, reordering, multiple control domains, measured network traces, and a process-separated asynchronous stack consistently support both diagnoses and the corresponding architecture-specific corrections.
\end{abstract}

\section{Introduction}
\suppressfloats[t]

\begin{figure}[t]
    \centering
    \includegraphics[width=\linewidth]{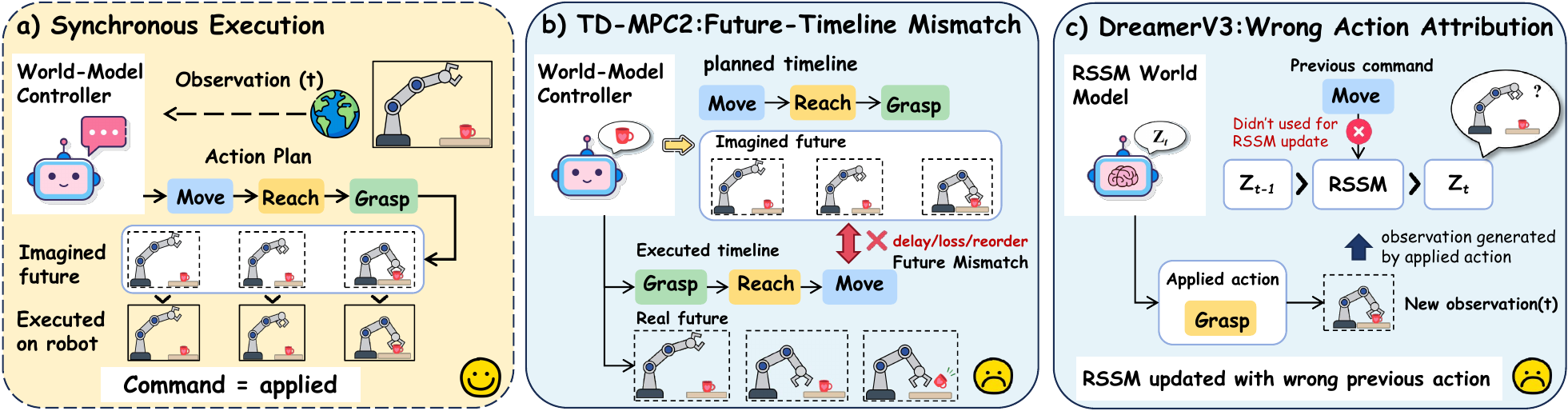}
    \caption{(a) Synchronous execution keeps commanded and applied actions aligned. Under asynchronous execution, (b) TD-MPC2 suffers from future-action timeline mismatch, while (c) DreamerV3 can suffer from previous-action attribution mismatch.}
    \label{fig:motivation}
    \vspace{-6pt}
\end{figure}

World models have become a central paradigm for model-based decision making, enabling agents to predict future outcomes, optimize policies through imagined trajectories, and plan actions online in learned latent dynamics~\citep{hansen2022tdmpc,hafner2023mastering,hansen2024tdmpc2,li2026causalworld,anand2026asyncworldmodels}.
These capabilities are particularly attractive for continuous control and robotics, where the controller must repeatedly predict how actions change the environment before acting~\citep{art:2023:daydreamer,brito2025wmreferences,li2025rewardfreewm}.
However, practical control systems are rarely perfectly synchronous: action computation, communication, buffering, and actuation may occur on different timelines~\citep{zhang2013networked,park2018wireless,wang2026remac}.
As a result, the action issued by a controller is not always the action that actually drives the corresponding environment transition~\citep{katsikopoulos2003markov,art:2021:rl-random-delays,art:2021:delay-aware-mbrl}.

Prior work on delayed and asynchronous control has followed several directions at different stages of the control loop. Early formulations recover a Markovian decision process by augmenting the state with delayed observations, pending actions, or action histories~\citep{katsikopoulos2003markov,walsh2009learning,Derman2021}. Later work uses learned dynamics or belief estimation to compensate for delayed interaction by propagating the effects of pending actions, reconstructing the current latent state, or incorporating multi-step action delay into model-based planning~\citep{art:2021:delay-aware-mbrl,art:2023:bpql,art:2024:dez}. More recent methods address stochastic or unobservable execution timing through delay-corrected learning, adaptive interaction models, or alternative future actions conditioned on uncertain execution times and packet loss~\citep{art:2021:rl-random-delays,art:2024:vdpo,wikman2026acda}. In parallel, recent robot-control systems consider the effects of slow policy inference on execution: RTC overlaps action generation with execution while preserving committed actions, REMAC addresses inconsistencies between current perception and concurrently executed action chunks, and FutureRTC predicts the state or observation available when new actions begin execution~\citep{black2025rtc,wang2026remac,jiang2026futurertc}.

\textbf{Execution-semantic gap.}
World-model control raises a different issue because actions serve both as outputs for execution and as explicit inputs to the learned predictive model. Existing delay-aware model-based methods mainly use learned dynamics to compensate for delayed interaction, while recent studies of asynchronous world models examine related misalignment on the observation side~\citep{art:2021:delay-aware-mbrl,anand2026asyncworldmodels}. Less well understood is how command--execution mismatch affects the action-conditioned computations inside a world-model controller when the action used by the model differs from the one that generates the corresponding environment transition. This effect depends on the architecture: planning-based controllers such as TD-MPC and TD-MPC2 roll future action sequences through learned latent dynamics, whereas recurrent world models such as DreamerV3 condition latent-state updates on previous actions~\citep{hansen2022tdmpc,hansen2024tdmpc2,hafner2023mastering}. We therefore hypothesize that the same execution mismatch affects these architectures differently, producing a future-action timeline mismatch in planning-based world models and a previous-action attribution mismatch in recurrent world models.

\textbf{Execution-consistent interfaces.}
Following latent-space MPC designs that roll candidate action sequences through learned dynamics, our ablations show that current applied-action or buffer information is insufficient for TD-MPC2; reliable planning instead depends on the future action timeline in the dynamics rollout (Figure~\ref{fig:finding_timeline}).
Inspired by packet-based communication and packetized predictive control in networked control systems, we introduce \textbf{Execution-Consistent Interfaces (ECI)}, a family of architecture-specific interfaces aligning world-model action inputs with actions realized by the execution system.
For planning-based world models, we instantiate ECI as \textbf{Future-Sequence Execution}, which transmits the planner-generated future action suffix to an actuator-side buffer, preserving the action timeline considered during planning.
For recurrent world models, we instantiate ECI as \textbf{Applied-Action Feedback}, which feeds the applied action into the recurrent state update, restoring the correct action--transition correspondence.
Both interfaces modify only the controller--execution interface and leave the pretrained world model unchanged during deployment.

\textbf{Key empirical findings.}
Our experiments reveal three consistent observations.
\textbf{(1)} Command--applied divergence degrades control even with a fixed planned command sequence, and reliable latent planning requires the correct future action timeline inside the dynamics rollout.
\textbf{(2)} Effective corrections differ across architectures: planning-based control benefits from preserving an updatable future action sequence, while recurrent control depends on timely action feedback consistent with training.
\textbf{(3)} These effects extend across TD-MPC/TD-MPC2, multiple control domains, measured network-delay traces, and a process-separated asynchronous execution stack.

\textbf{Contributions}.Our main contributions are threefold. 
\textbf{(1) } We identify execution-semantic mismatch as a failure mode of world-model control and show that it manifests as a future-action timeline mismatch in planning-based TD-MPC2 and as a previous-action attribution mismatch in recurrent DreamerV3. 
\textbf{(2) } Guided by this diagnosis, we instantiate execution-consistent interfaces that preserve the planner's future action timeline for TD-MPC2 and feed back the actually applied action for DreamerV3, without modifying the underlying pretrained world models. 
\textbf{(3) } We validate these mechanisms across controller architectures, control tasks, delay, packet loss and reordering, measured network traces, and a process-separated asynchronous execution stack.

\section{Why Do World-Model Controllers Fail Under Asynchronous Execution?}
\label{sec:findings}

In this section, we conduct a series of controlled empirical analyses to understand why world-model controllers degrade under asynchronous execution. Rather than treating delay, packet loss, or reordering merely as external disturbances, we focus on how these execution effects alter the action information consumed by the world model.

\paragraph{Preliminaries of asynchronous world-model control.}
At each control step $t$, the controller produces a command action $u_t^c$, while the execution system applies an applied action $u_t^a$ to the environment. Under synchronous execution, $u_t^c=u_t^a$; under asynchronous execution, delay, packet loss, reordering, or buffering may instead lead to $u_t^c\neq u_t^a$. More generally, the applied action depends on the history of issued commands and the execution state:

\begin{equation}
    u_t^a = g(u_{\leq t}^c, e_t),
\end{equation}
where $e_t$ denotes execution-side information such as packet arrivals, delays, and buffer contents. This distinction is important because world models are action-conditioned: planner-based controllers, such as TD-MPC and TD-MPC2, use future action sequences to roll latent states forward, whereas recurrent world models, such as DreamerV3, condition recurrent state updates on the previous action and the newly observed transition. As a result, asynchronous execution can induce different forms of action-semantic mismatch depending on where each architecture consumes action information. \textbf{Q1:} How does command--applied action divergence degrade planning-based world-model control, and what execution information is needed to correct it? \textbf{Q2:} Does execution-semantic mismatch manifest differently across world-model architectures?


\subsection{Finding 1: Planning-Based World Models Need the Correct Future Action Timeline}
\label{sec:finding_timeline}

\begin{wrapfigure}[13]{r}{0.35\linewidth}
    \vspace{-10pt}
    \centering
    \includegraphics[width=\linewidth]{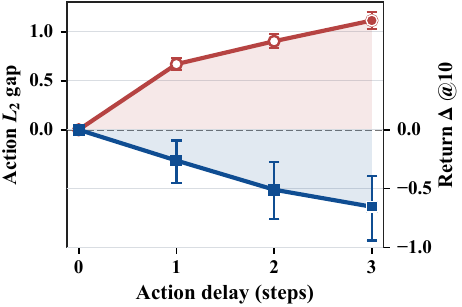}
    \vspace{-10pt}
    \caption{Effect of execution delay under fixed commands. Error bars denote 95\% confidence intervals.}
    \label{fig:finding1_divergence}
    \vspace{-20pt}
\end{wrapfigure}

We first isolate asynchronous execution from planner behavior by replaying the same planned command sequence from matched environment states while changing only the action-delivery delay. Figure~\ref{fig:finding1_divergence} shows that greater delay increases command--applied divergence and reduces short-horizon return, even though the planner makes the same decisions. Thus, the execution process itself can degrade control, suggesting that TD-MPC2 may be missing information about how its commands will actually be realized.

To identify this missing information, we progressively provide TD-MPC2 with the current applied action, the actuator buffer, or the exact future applied-action sequence. The first two describe only the current execution state, whereas the last specifies the action timeline that will actually drive the environment.

\begin{figure}[H]
    \centering
    \includegraphics[width=\linewidth]{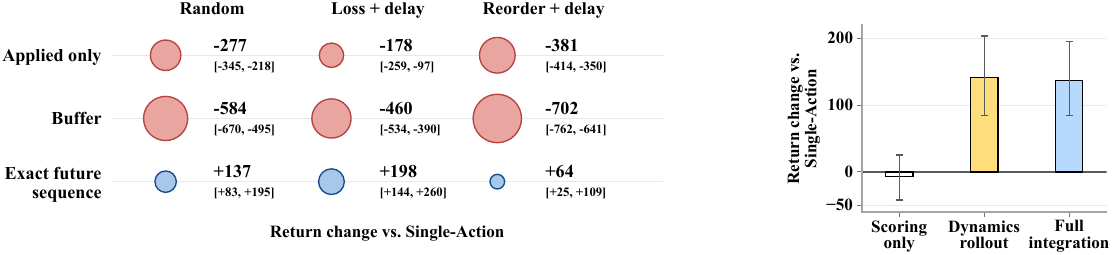}
    \caption{TD-MPC2 execution-information ablations. Left: information type across perturbations. Right: insertion point. Brackets and error bars denote 95\% CIs.}
    \label{fig:finding_timeline}
\end{figure}

Figure~\ref{fig:finding_timeline} shows that current applied-action or buffer information is insufficient, while the exact future sequence improves control only when it enters the latent dynamics rollout; using the same information for trajectory rescoring alone provides no benefit. This identifies the failure more precisely: TD-MPC2 predicts future states under one action sequence, while asynchronous execution may realize another. We refer to this as a \emph{future-action timeline mismatch}. Reliable planning therefore requires the action timeline used inside the world-model rollout to remain consistent with the one realized by the execution system.


\subsection{Finding 2: Execution-Semantic Mismatch Manifests Differently across World-Model Architectures}
\label{sec:finding_architecture}

We next ask whether execution-semantic mismatch also appears in world-model architectures that consume actions differently. Unlike TD-MPC2, which uses future actions during latent rollout, DreamerV3 conditions its recurrent state update on the previous action and the new observation. Under asynchronous execution, the environment may evolve under one applied action while the recurrent model attributes the transition to a different issued command.

\begin{wrapfigure}[11]{r}{0.53\linewidth}
    \vspace{-8pt}
    \centering
    \includegraphics[width=\linewidth]{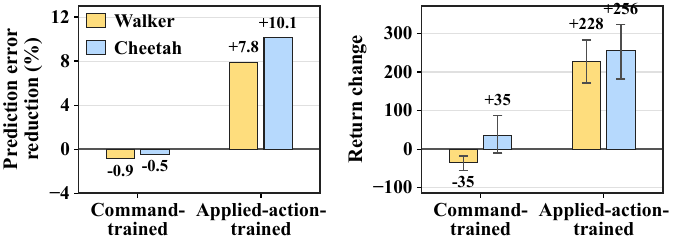}
    \vspace{-6pt}
    \caption{DreamerV3 feedback alignment under random delay.}
    \label{fig:finding2_prediction_control}
    \vspace{-8pt}
\end{wrapfigure}

To isolate this effect, we vary only the previous-action input to DreamerV3's recurrent update while holding the observed trajectory and forecast action fixed. We evaluate both command-trained and applied-action-trained checkpoints, allowing us to test whether the benefit of applied-action feedback depends on the action semantics learned during training.

Figure~\ref{fig:finding2_prediction_control} links prediction and control: the left panel reports prediction-error reduction, while the right reports return change with 95\% CIs. Applied-action feedback improves both measures for applied-action-trained checkpoints, but is not uniformly beneficial for command-trained models. This indicates that the recurrent-state update is sensitive not only to which action is supplied, but also to the action semantics learned during training. Together with Finding~1, this reveals two distinct failure modes: a future-action timeline mismatch in TD-MPC2 and a previous-action attribution mismatch in DreamerV3. Execution corrections should therefore target where each architecture consumes action information and remain consistent with its learned action semantics.

\section{Execution-Consistent Interfaces}
\label{sec:method}

\begin{figure}[t]
\centering
\includegraphics[width=\linewidth]{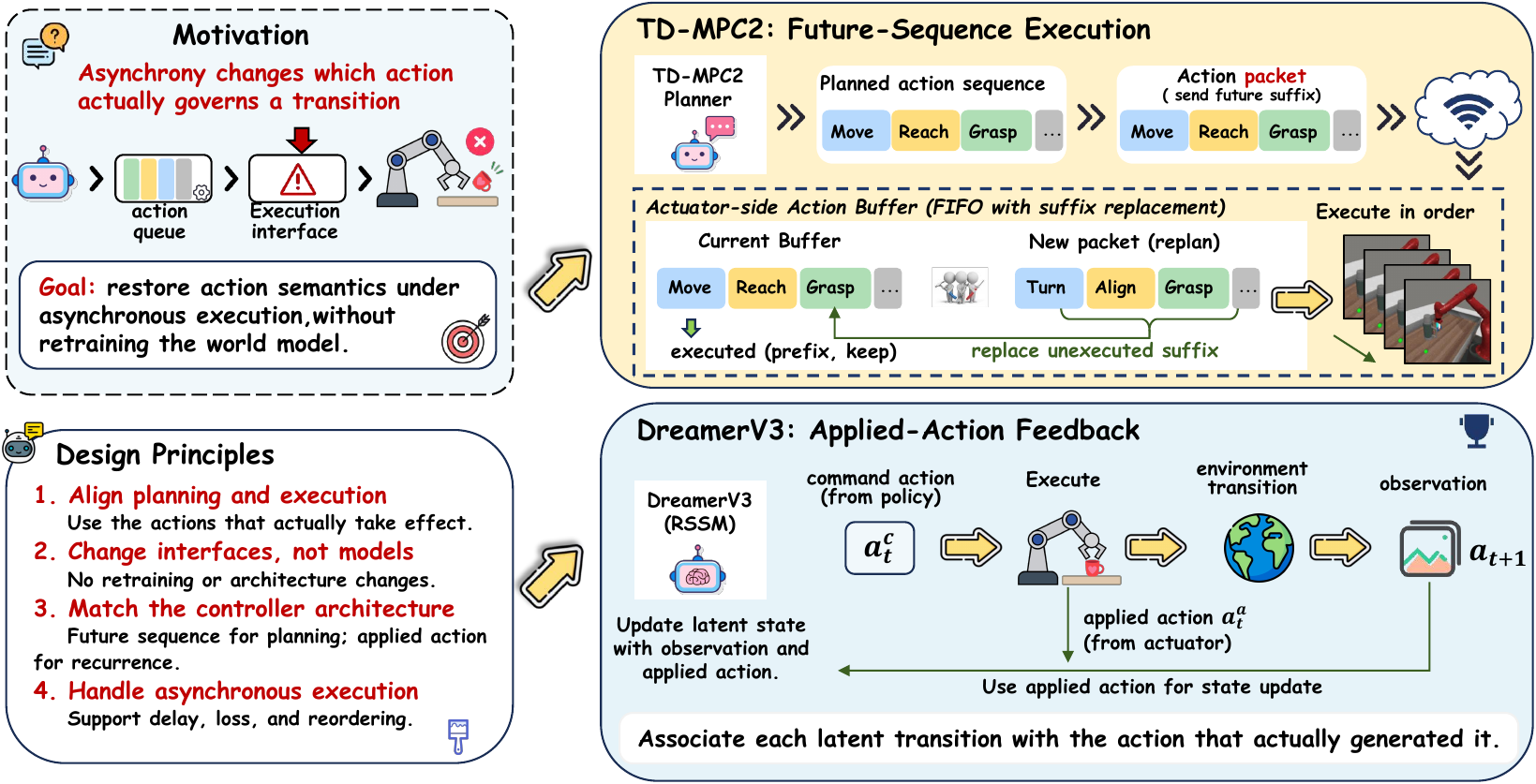}
\caption{Overview of our execution-consistent world-model control framework.}
\label{fig:framework}
\end{figure}

After identifying where asynchronous execution breaks the action semantics of
world-model controllers, inspired by packetized transmission and receiver-side
buffering in computer networks~\citep{kurose2021computer}, we introduce \textbf{E}xecution-\textbf{C}onsistent \textbf{I}nterfaces (ECI)
, which restore the action information required by each architecture
without modifying the pretrained world model. As illustrated in Figure~\ref{fig:framework}, our approach consists of future-sequence execution for planning-based world models (Section~\ref{sec:method_tdmpc}) and applied-action feedback for recurrent world models (Section~\ref{sec:method_dreamer}).

\subsection{Future-Sequence Execution for Planning-Based World Models}
\label{sec:method_tdmpc}

\textbf{Future-sequence transmission.}
At each control step $t$, TD-MPC2 produces a future action sequence over horizon $H$:
\begin{equation}
U_t^c =
\left(
u_{t|t}^c,
u_{t+1|t}^c,
\ldots,
u_{t+H-1|t}^c
\right),
\end{equation}
where $u_{t+h|t}^c$ denotes the action planned at time $t$ for time $t+h$. Under standard receding-horizon execution, only the current action $u_{t|t}^c$ is sent to the actuator. Under asynchronous execution, the next command can arrive too late because of transmission delay, be missing because of packet loss, or arrive out of order. The actuator then continues using actions from an earlier plan. To preserve the planner's future control information, we transmit the entire planner-derived sequence rather than only its first action. Each control packet contains the current action together with the remaining future actions and a generation index, providing the execution system with a ready-to-use continuation of the latest plan when subsequent commands are delayed, lost, or reordered.

\textbf{Replaceable action buffer.}
Each transmitted sequence is associated with its planning time and stored at the actuator. Let $\tau_t$ denote the arrival time of the sequence generated at time $t$. At execution step $k$, the actuator uses the newest valid sequence that has already arrived and selects the action corresponding to the current step:
\begin{equation}
j_k
=
\max\{t \mid \tau_t \leq k\},
\qquad
u_k^{a}
=
\begin{cases}
u_{k|j_k}^{c}, & 0 \leq k-j_k < H,\\
u_k^{\mathrm{fb}}, & \text{otherwise},
\end{cases}
\label{eq:buffer_execution}
\end{equation}
where $u_k^{\mathrm{fb}}$ denotes the fallback action when no valid
planned action remains; we use a zero action in our experiments. When a newer sequence arrives, its unexecuted portion replaces the remaining actions from the previous sequence. Sequences that arrive after a newer plan has already been accepted are discarded, preventing stale plans from overwriting the current buffer.

\textbf{Execution-consistent planning.}
The proposed interface does not predict future communication delays or introduce an additional control policy. Instead, it preserves future control information that TD-MPC2 has already produced. When a newly replanned command is unavailable, the actuator continues along the remaining actions of the newest available plan rather than immediately switching to an unrelated fallback action. This reduces the discrepancy between the future action sequence evaluated during world-model rollout and the sequence available for execution, directly addressing the future action timeline mismatch identified in Finding~1.

\subsection{Applied-Action Feedback for Recurrent World Models}
\label{sec:method_dreamer}

\textbf{Applied-action feedback.}
DreamerV3 does not explicitly plan a future action sequence during online execution. Instead, its RSSM updates the recurrent state using the previous latent state, the previous action, and the current observation:
\begin{equation}
s_t = f_{\mathrm{RSSM}}(s_{t-1}, a_{t-1}, o_t).
\end{equation}
Under synchronous execution, the issued command is also the action that generates the transition leading to $o_t$. Under asynchronous execution, however, the environment may actually evolve under a different applied action. If the RSSM still conditions on the issued command, it can associate the observed transition with the wrong action. We therefore replace the command action with the action that was actually applied:
\begin{equation}
s_t = f_{\mathrm{RSSM}}(s_{t-1}, u_{t-1}^{a}, o_t).
\end{equation}

\textbf{Execution receipt.}
To obtain the applied action without accessing simulator-internal information, the actuator returns an execution receipt together with the subsequent observation. The receipt identifies which action generated the transition, including its source packet and its position within that packet. It can either carry the applied-action vector directly or provide an identifier that allows the controller to recover the corresponding action from its local command history. Because the receipt is piggybacked on the observation, no additional communication round trip is required.

\textbf{Correct action attribution.}
The purpose of this interface is not to provide DreamerV3 with future execution information, but to ensure that each observed transition is paired with the action that actually produced it. When the execution receipt arrives in time, the RSSM can update its recurrent state using the correct action, directly addressing the previous action attribution mismatch identified in Finding~2. The intervention changes only the information exchanged between the controller and the execution system, while leaving the pretrained world model and actor unchanged.

\begin{table*}[t]
\centering
\caption{Normalized TD-MPC2 return under different execution interfaces, averaged equally across tasks and normalized by matched clean \textsc{Single-Action} return. Bold and underlined values mark the best and second-best TD-MPC2 results excluding the Oracle.}
\label{tab:tdmpc2_main}
\small
\setlength{\tabcolsep}{7pt}
\begin{tabular}{@{}lcccccc@{}}
\toprule
Method & Clean & Random & Bursty & Packet loss & Reordering & Adverse avg. \\
\midrule
\rowcolor{TableBlue}[0pt][0pt]
\multicolumn{7}{@{}l@{}}{\textbf{Execution baselines}} \\
\textsc{Single-Action}        & 1.000 & 0.305 & 0.215 & 0.232 & 0.358 & 0.278 \\
\textsc{Hold-Last}            & 1.000 & 0.325 & 0.205 & 0.243 & 0.379 & 0.288 \\
\textsc{Fixed Chunk}          & 1.000 & 0.286 & 0.211 & 0.268 & 0.305 & 0.268 \\
\textsc{Receding Buffer}      & 1.000 & \textbf{0.783} & \textbf{0.605} & \underline{0.537} & \textbf{0.942} & \underline{0.717} \\
\midrule
\textbf{\textsc{Future-Sequence} (ours)}
& 1.000
& \textbf{0.783}
& \textbf{0.605}
& \textbf{0.601}
& \textbf{0.942}
& \textbf{0.733} \\
\midrule
\rowcolor{TablePeach}[0pt][0pt]
\multicolumn{7}{@{}l@{}}{\textbf{External methods}} \\
DATS & 0.266 & 0.197 & 0.182 & 0.142 & 0.173 & 0.174 \\
ACDA & 0.407 & 0.458 & 0.430 & 0.422 & 0.433 & 0.436 \\
\midrule
\rowcolor{TableGray}[0pt][0pt]
\multicolumn{7}{@{}l@{}}{\textbf{Oracle diagnostic}} \\
\textsc{Future-Action Oracle} & 1.000 & 0.880 & 0.834 & 0.682 & 0.892 & 0.822 \\
\bottomrule
\end{tabular}
\end{table*}

\section{Experiments}
\label{sec:experiments}

\subsection{Setup}
\label{sec:exp_setup}

\paragraph{Controllers and protocol.}
We study two structurally different world-model controllers, TD-MPC2 and DreamerV3. Unless explicitly studying delay-aware retraining, all pretrained models remain frozen during evaluation. We consider clean execution, fixed and random action delays, packet loss, and packet reordering, and further validate selected settings using measured network-delay traces and a process-separated asynchronous execution stack.

\paragraph{Benchmarks and metrics.}
Our primary experiments use the DeepMind Control Suite~\citep{tassa2018deepmind}: TD-MPC2 is evaluated on Dog, Walker, and Hopper, while DreamerV3 is evaluated on Walker, Cheetah, and Cup. Cross-domain experiments further cover MetaWorld~\citep{yu2020metaworld}, ManiSkill2~\citep{gu2023maniskill2}, and MyoSuite~\citep{caggiano2022myosuite}. We report performance using three key metrics: \textbf{episodic return} for overall control performance, \textbf{task success rate} for tasks with explicit completion criteria, and \textbf{command--applied action divergence} for quantifying execution-semantic mismatch. Deployment-oriented experiments additionally report per-step execution overhead and communication statistics. Primary results are reported with 95\% confidence intervals.

\paragraph{Baselines.}
For TD-MPC2, we compare four execution-side baselines: \textsc{Single-Action} sends only the current planned action; \textsc{Hold-Last} repeats the most recently available action when a new command is unavailable; \textsc{Fixed Chunk} transmits and executes a fixed multi-action chunk; and \textsc{Receding Buffer} maintains a buffered action sequence that is refreshed as new plans arrive. Our method, \textsc{Future-Sequence}, preserves the planner-generated future action sequence at the execution side, while a privileged \textsc{Future-Action Oracle} uses future applied actions only for diagnosis. We also compare with delay-aware learning methods, DATS~\citep{art:2021:delay-aware-mbrl} and ACDA~\citep{wikman2026acda}, which adapt the controller during delayed training. For DreamerV3, \textsc{CMD} conditions the recurrent update on the issued command, \textsc{RCP} repeatedly executes the current command, and our \textsc{APP} instead feeds back the actually applied action. \textsc{ECT-Full} and \textsc{ECT-ID} implement APP in the process-separated stack using actuator-originated execution receipts.

\begin{figure}[t]
\centering
\includegraphics[width=\linewidth]{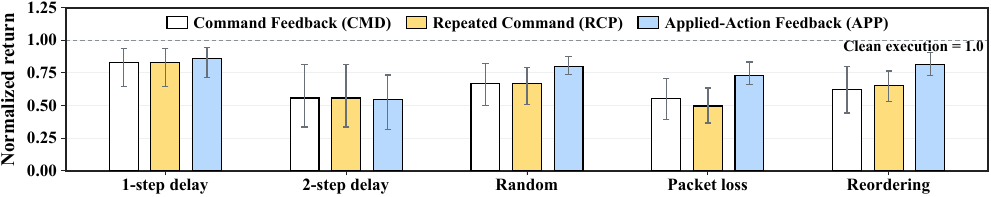}
\caption{DreamerV3 under asynchronous execution. Each task is normalized by its mean clean-execution return under CMD and averaged equally across tasks; error bars denote 95\% CIs.}

\label{fig:dreamer_main}
\end{figure}

\subsection{Experimental Results}
\label{sec:exp_results}

\begin{wraptable}[14]{l}{0.40\linewidth}
    \vspace{-20pt}
    
    \centering
    \setlength{\belowcaptionskip}{5pt}
    \caption{Future-Sequence return change on Walker.}
    \label{tab:delay_training_difference}
    \footnotesize
    
    \setlength{\tabcolsep}{2pt}
    \renewcommand{\arraystretch}{1.08}
    \begin{tabularx}{\linewidth}{@{}Xrrr@{}}
    \toprule
    \rowcolor{TableBlue}
    \textbf{Evaluation} & \textbf{Clean} & \textbf{Fixed-2} & \textbf{Random} \\
    \midrule
    Clean           &   +9 &  +10 &   +3 \\
    \addlinespace[1pt]
    Fixed-1         & +220 & -324 & -337 \\
    Fixed-2         &  -15 & -473 & -274 \\
    Fixed-3         & -110 & -343 & -207 \\
    \addlinespace[1pt]
    Random          & +500 &  -28 & +275 \\
    Loss + delay    & +532 &  -19 & +270 \\
    Reorder + delay & +525 &  -32 & +294 \\
    \midrule
    \rowcolor{TableGray}
    \textbf{Adverse avg.} & \textbf{+276} & \textbf{-203} & \textbf{+3} \\
    \bottomrule
    \end{tabularx}
    \vspace{-6pt}
\end{wraptable}

\paragraph{Cross-architecture results.}
Table~\ref{tab:tdmpc2_main} and Figure~\ref{fig:dreamer_main} summarize the main results on frozen TD-MPC2 and DreamerV3 controllers. For TD-MPC2, Future-Sequence preserves clean performance, substantially improves robustness across all adverse execution conditions, and achieves the highest reported adverse-condition average among the TD-MPC2 execution interfaces excluding the Oracle. It outperforms Single-Action, Hold-Last, and Fixed Chunk across all reported perturbations, matches Receding Buffer at the reported precision under random delay, bursty delay, and reordering, and achieves higher return under packet loss. For DreamerV3, Applied-Action Feedback provides the clearest gains under random delay, packet loss, and reordering, whereas Repeated Command offers little consistent benefit and fixed-delay improvements are less uniform. Together, these results support architecture-specific execution corrections: preserving the planner-generated future action sequence for planning-based world models, and correcting action attribution for recurrent world models. Additional numerical comparisons for DreamerV3 are provided in Appendix~\ref{sec:dreamer_results}. Figure~\ref{fig:qualitative_rollout_case} visualizes the difference within a paired Walker-Walk rollout under packet reordering. Both methods use the same checkpoint, initial state, environment seed, and transport seed. Single-Action loses balance early and attains a return of 397.4, whereas Future-Sequence remains upright throughout the episode and reaches 889.2. The replayed frames and body-height traces show how preserving a coherent future action sequence turns the aggregate return gain into sustained physical stability.

\paragraph{Generalization results.}
We further test Future-Sequence beyond the primary TD-MPC2 setting. It produces consistent gains on the original TD-MPC controller, across held-out MetaWorld, ManiSkill2, and MyoSuite tasks, and under measured network-delay profiles. These results show that the benefit is not restricted to TD-MPC2, a single task domain, or synthetically generated delay.

\begin{figure}[tbp]
\centering
\includegraphics[width=\linewidth]{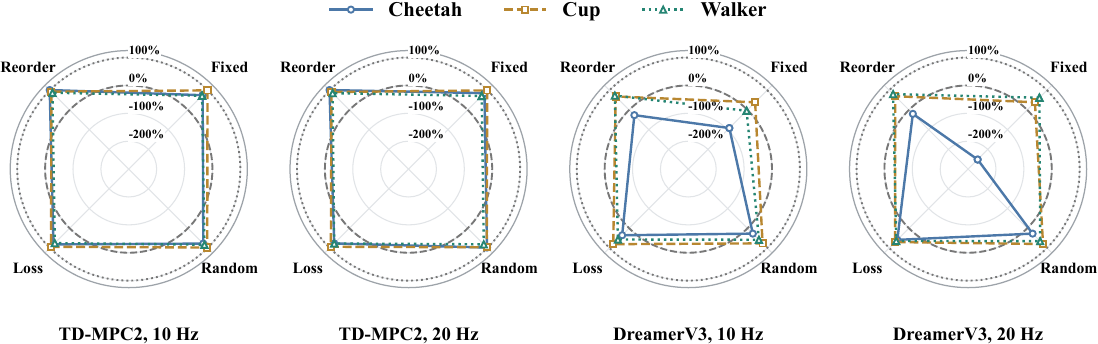}
\caption{Lost return recovered in the asynchronous stack. From left: TD-MPC2 at 10/20 Hz, then DreamerV3 at 10/20 Hz. Dashed and dotted rings mark 0\% and 100\% recovery. All panels share a radial scale starting at -300\%; negative values indicate further degradation.}
\label{fig:async_stack_recovery}
\end{figure}

\paragraph{Deployment results.}
Figure~\ref{fig:async_stack_recovery} shows that the corrections remain effective when the controller and actuator run in separate processes communicating over UDP. Under injected delay, loss, and reordering, Future-Sequence recovers TD-MPC2 performance across the evaluated conditions, while Applied-Action Feedback improves DreamerV3 in most conditions. The required action feedback can be supplied directly by the actuator: \textsc{ECT-Full} returns the applied-action vector, whereas \textsc{ECT-ID} identifies its source packet and position. Both recover the exact applied action across 30,000 tested transitions, demonstrating that action attribution can be implemented through execution receipts rather than privileged simulator access.

\subsection{Analysis}
\label{sec:analysis}

\paragraph{Training--interface interaction analysis.}
A natural question is whether Future-Sequence remains useful when the controller is already trained under delayed execution. Table~\ref{tab:delay_training_difference} reveals a strong interaction. Clean-trained checkpoints benefit broadly, whereas combining Future-Sequence with Fixed-2 training degrades every adverse condition. Random-delay training is mixed: it improves stochastic delay, loss, and reordering but degrades fixed-delay conditions, leaving the adverse average nearly unchanged. This pattern shows that delay-aware training and execution-side correction are not automatically additive. The most plausible explanation is that delay-aware training changes the implicit action semantics of the controller, so an execution interface designed for a clean-trained controller can become mismatched or compensate twice. Absolute returns are reported in Appendix~\ref{sec:delay_training_results}.

\begin{wrapfigure}[15]{r}{0.40\linewidth}
    \vspace{-8pt}
    \centering
    \includegraphics[width=\linewidth]{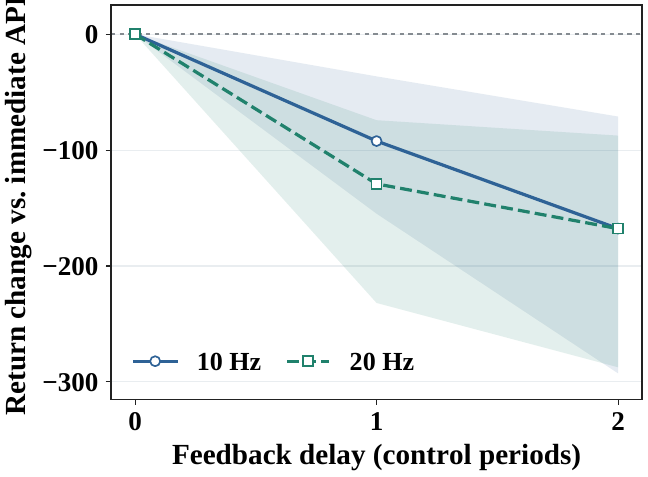}
    \vspace{-5pt}
    \caption{Feedback latency under random delay. Shading: 95\% CI.}
    \label{fig:feedback_latency}
    \vspace{-8pt}
\end{wrapfigure}


\paragraph{Feedback-latency analysis.}
Figure~\ref{fig:feedback_latency} examines how feedback timeliness affects Applied-Action Feedback. At both evaluated control rates, immediate feedback yields the highest aggregate return, while one- and two-period feedback delays progressively reduce performance. Thus, access to applied actions alone is not sufficient: their temporal correspondence with the observed transitions also matters. This complements the action-semantics analysis in Section~\ref{sec:finding_architecture}, identifying timely, transition-aligned feedback as an important requirement for effective recurrent-state correction.

\paragraph{Execution-buffer analysis.}
Table~\ref{tab:tdmpc2_main} shows that preserving and updating a future action sequence outperforms transmitting isolated actions or executing fixed chunks. Both Future-Sequence and Receding Buffer substantially improve asynchronous control. They match at reported precision under random delay, bursty delay, and reordering, while Future-Sequence performs better under packet loss. Their implementations differ in updating the remaining sequence with incoming plans and handling execution after buffer exhaustion (Appendix~\ref{sec:baseline_details}). Together with Figure~\ref{fig:finding_timeline}, these results highlight the central role of retaining the planner's future action timeline, with buffer management further shaping robustness.

\begin{figure}[t]
    \centering
    \includegraphics[width=\linewidth]{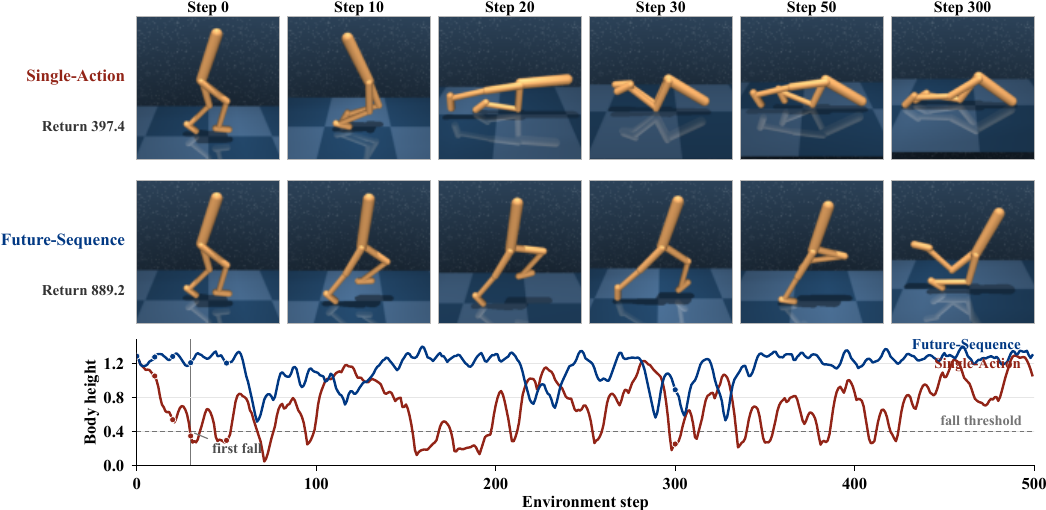}
    \caption{Paired Walker-Walk rollout under packet reordering.}
    \label{fig:qualitative_rollout_case}
\end{figure}

\section{Related Work}
\label{sec:related_work}

\subsection{Delayed reinforcement learning.}
Delayed reinforcement learning studies decision making when observations or actions are delayed. Methods fall into three categories. The first augments state with pending actions or action histories to recover the Markov property \citep{walsh2009learning,Derman2021}. The second exploits delayed dynamics through state prediction or model-based reasoning, including real-time actor--critic, delay-aware model-based reinforcement learning, and belief-projection methods \citep{art:2021:delay-aware-mbrl,art:2023:bpql}. The third addresses stochastic or partially observable delays through delay-corrected learning, latent planning, or adaptive interaction models \citep{art:2021:rl-random-delays,art:2024:vdpo,art:2024:dez,wikman2026acda}. Overall, this work progressed from representing delayed interactions to inferring decision-relevant states and actions under uncertain execution timing. Its primary objective is effective decision making under delay, not understanding how execution timing alters action-conditioned computations within a world-model controller.

\subsection{World-model control under asynchronous execution.}
World-model controllers use action-conditioned dynamics for prediction, policy optimization, and planning, including recurrent latent models such as PlaNet and Dreamer \citep{hafner2023mastering}, tree-search methods including MuZero and EfficientZero \citep{schrittwieser2020mastering,Ye2021}, and latent trajectory optimization in TD-MPC and TD-MPC2 \citep{hansen2022tdmpc,hansen2024tdmpc2}. World models handle delayed interaction via future-state prediction, delayed-observation imagination, and delay-aware planning \citep{wikman2026acda}. These works mainly use world models to compensate for delay rather than study how asynchronous execution can invalidate assumed action semantics. Recent robot-learning methods emphasize prediction--execution alignment: RTC supports asynchronous action-chunk generation and execution, REMAC addresses inconsistencies between perception and concurrent actions, and FutureRTC studies prediction--execution misalignment \citep{black2025rtc,wang2026remac,jiang2026futurertc}; related work studies asynchronous sensor observations in world models \citep{anand2026asyncworldmodels}. We therefore study the action side of world-model control under asynchronous execution and show how command--applied divergence induces future-action timeline mismatch in planning-based models and previous-action attribution mismatch in recurrent models.

\section{Conclusion}
In this paper, we identify how asynchronous execution disrupts action-conditioned computation in world-model controllers. Controlled interventions reveal a future-action timeline mismatch in TD-MPC2 and a training-dependent previous-action attribution mismatch in DreamerV3. Guided by these findings, execution-consistent interfaces improve control across tasks and asynchronous execution conditions. Our results highlight the importance of aligning execution-time action information with both model computation and learned action semantics.






\bibliography{iclr2027_conference}
\bibliographystyle{iclr2027_conference}

\appendix
\section{DreamerV3 Numerical Results}
\label{sec:dreamer_results}

Table~\ref{tab:dreamer_main} complements Figure~\ref{fig:dreamer_main} with the clean-normalized return improvement over Command Feedback. For each task, we divide the return difference by its mean clean-execution return under Command Feedback, then average equally across Walker, Cheetah, and Cup.

\begin{table}[htbp]
\centering
\caption{DreamerV3 clean-normalized return improvement over Command Feedback. Means use ten evaluation seeds per task and equal task weights.}
\label{tab:dreamer_main}
\small
\setlength{\tabcolsep}{3.5pt}
\begin{tabular}{lcccccc}
\toprule
Method & Clean & 1-step delay & 2-step delay & Random & Packet loss & Reordering \\
\midrule
\textsc{Command Feedback}
& 0.000 & 0.000 & 0.000 & 0.000 & 0.000 & 0.000 \\
\textsc{Repeated Command}
& 0.000 & 0.000 & 0.000 & -0.001 & -0.055 & 0.028 \\
\midrule
\textsc{Applied-Action Feedback}
& 0.000 & 0.032 & -0.013 & \textbf{0.131} & \textbf{0.180} & \textbf{0.192} \\
\bottomrule
\end{tabular}
\end{table}

\section{Delay-Aware Training Numerical Results}
\label{sec:delay_training_results}

Table~\ref{tab:delay_learning} provides the absolute returns underlying Table~\ref{tab:delay_training_difference}.

\begin{table}[H]
\centering
\caption{
Walker episodic return under different training and execution conditions. Entries average five independently trained checkpoints and ten evaluation seeds per checkpoint. F1 to F3 denote fixed delays, and Adv. denotes the mean over all non-clean conditions.
}
\label{tab:delay_learning}
\scriptsize
\setlength{\tabcolsep}{2.6pt}
\begin{tabular}{llrrrrrrrr}
\toprule
Training & Execution & Clean & F1 & F2 & F3 & Random & Loss & Reorder & Adv. \\
\midrule
Clean & \textsc{Single-Action}
& 965 & 603 & 306 & 194 & 290 & 205 & 365 & 327 \\

Clean & \textsc{Future-Sequence}
& \textbf{974} & \textbf{824} & 292 & 84
& \textbf{790} & \textbf{737} & \textbf{890} & \textbf{603} \\

\midrule
Fixed-2 & \textsc{Single-Action}
& 118 & 373 & 512 & \textbf{379} & 89 & 76 & 104 & 256 \\

Fixed-2 & \textsc{Future-Sequence}
& 129 & 49 & 39 & 36 & 61 & 58 & 72 & 52 \\

\midrule
Random 0--3 & \textsc{Single-Action}
& 896 & 812 & \textbf{558} & 337 & 313 & 248 & 384 & 442 \\

Random 0--3 & \textsc{Future-Sequence}
& 899 & 475 & 284 & 130 & 588 & 518 & 678 & 445 \\
\bottomrule
\end{tabular}
\end{table}

\section{Implementation Details}
\label{sec:implementation_details}

\paragraph{Frozen evaluation.}
All execution-interface comparisons keep the controller parameters fixed. For each checkpoint, methods are paired using the same environment and transport seeds. At the beginning of each episode, the execution queue, held action, and controller state are reset. Controller training and execution-interface evaluation are conducted as separate experiments.

\paragraph{TD-MPC2 diagnostic planner.}
The execution-information and insertion-point ablations use a planning horizon of five, 64 candidate trajectories, eight policy-prior trajectories, and three base planning iterations. Following the original TD-MPC2 implementation, two additional iterations are used when the action dimension is at least 20. Returns are evaluated over 500-step episodes. During diagnostic evaluation, the planner's random-number state is preserved so that additional score queries do not alter subsequent planning samples.

\paragraph{DreamerV3 action alignment.}
The action supplied to the recurrent update is aligned with the transition that produces the current observation. Command Feedback uses the command issued for that transition, whereas Applied-Action Feedback uses the action actually applied to the environment. This action is distinct from the next command generated after observing the transition. The main DreamerV3 evaluation uses action repeat one and ten evaluation seeds per task and condition.

\section{Baseline Definitions}
\label{sec:baseline_details}

\paragraph{TD-MPC2 interfaces.}
\textsc{Single-Action} transmits only the first planned action. \textsc{Hold-Last} reuses the most recently available action when no new command arrives. \textsc{Fixed Chunk} accepts a new action sequence only after its active queue is exhausted. \textsc{Receding Buffer} accepts newer plans, discards actions whose scheduled execution times have passed, and replaces the remaining queue; after exhaustion, it holds the last applied action. Our \textsc{Future-Sequence} interface stores actions according to their intended execution slots, updates overlapping future slots when newer plans arrive, and rejects stale plans. The registered diagnostic implementation uses a zero-action fallback when no valid planned action remains.

\paragraph{Future-Action Oracle.}
The \textsc{Future-Action Oracle} evaluates candidate actions using an isolated copy of the network and actuator state, including future packet-delivery outcomes. The resulting applied-action sequence is then supplied to the world-model rollout. This Oracle is a privileged diagnostic condition rather than a deployable interface or a guaranteed upper bound on return.

\paragraph{DreamerV3 interfaces.}
Command Feedback (\textsc{CMD}) supplies the issued command to the recurrent update. Repeated Command (\textsc{RCP}) transmits five copies of the current command rather than a predicted future action sequence. Our Applied-Action Feedback (\textsc{APP}) instead supplies the actual previous action. In the process-separated implementation, \textsc{ECT-Full} returns the applied-action vector, whereas \textsc{ECT-ID} returns its source packet and position, allowing the controller to reconstruct the action from its command history.

\subsection{External Learning-Based Baselines}
\label{sec:external_baseline_details}

\paragraph{DATS.}
We reproduce DATS~\citep{art:2021:delay-aware-mbrl} using an ensemble of five probabilistic dynamics models, each with four hidden layers of width 200. The models predict transition means and variances and are trained with a learning rate of $10^{-3}$, batch size 256, and five epochs per update. Planning uses CEM with horizon 30, 500 candidates, 50 elites, and five optimization iterations. Following DATS, pending actions are prepended to each candidate suffix so that planning accounts for actions already committed to execution. Training uses a fixed two-step action delay. For Walker, we use the analytic task reward; for Dog and Hopper, the learned model additionally predicts the observed reward.

\paragraph{ACDA.}
We adapt the released ACDA implementation~\citep{wikman2026acda} to the DMC task interface and our runtime. ACDA uses a delay-indexed action matrix with horizon four, policy hidden layers of width 256, predictive-model hidden and recurrent dimensions of 256 and 384, and a model window of 16. The batch size is 256, replay capacity is $10^6$, and the predictive-model learning rate is $10^{-4}$. Delay memorization and model-output clamping are enabled. At execution time, ACDA produces its native delay-conditioned action matrix rather than a TD-MPC2 future sequence.

\paragraph{Evaluation protocol.}
DATS and ACDA train their own predictive models and controllers, whereas our Future-Sequence interface operates on a pretrained TD-MPC2 controller without modifying its parameters. We therefore evaluate each method as its complete controller--execution system. Checkpoints remain fixed during evaluation, and stochastic evaluations use ten environment seeds per checkpoint and condition with 500 control decisions per episode.

\section{Ablation Protocols}
\label{sec:ablation_details}

\paragraph{Fixed-command intervention.}
Figure~\ref{fig:finding1_divergence} replays the same planned command sequence from matched environment states while varying action-delivery delay from zero to three steps. We measure command--applied action divergence and the change in ten-step return relative to zero-delay execution.

\paragraph{Execution-information ablation.}
Figure~\ref{fig:finding_timeline} compares three forms of execution information supplied during planning. \textsc{Applied} replaces only the first rollout action with the currently applied action. \textsc{Buffer} replaces rollout actions for time slots covered by the current actuator buffer while leaving the remaining candidate actions unchanged. \textsc{Exact Future Sequence} maps each candidate sequence to the actions produced by a copied execution system. The actual execution mechanism is held fixed across all information conditions. Each plotted contrast uses 20 paired episodes per perturbation.

\paragraph{Insertion-point ablation.}
Under random delay, \textsc{Scoring Only} supplies the mapped actions to the reward predictor while retaining command actions in the latent dynamics. \textsc{Dynamics Rollout} supplies mapped actions to the latent dynamics while retaining command actions in the reward predictor. \textsc{Full Integration} uses mapped actions in both components. All variants use the same terminal policy and value-estimation rule and are evaluated with 20 paired episodes. \textsc{Full Integration} therefore changes only where execution information enters the planner and does not involve additional model training.

\paragraph{Prediction and control alignment.}
Figure~\ref{fig:finding2_prediction_control} evaluates Walker and Cheetah using command-trained and applied-action-trained checkpoints. For the prediction diagnostic, the observation trajectory and forecast action are fixed while only the previous-action input to the recurrent update is changed. Each checkpoint--task condition uses one held-out episode containing 950 evaluated prediction steps. Prediction improvement is defined as
\[
100\frac{M_{\mathrm{CMD}}-M_{\mathrm{APP}}}{M_{\mathrm{CMD}}},
\]
where $M$ denotes observation-prediction MSE. Control improvement is evaluated separately using ten paired closed-loop episodes. Its 95\% confidence interval is estimated with 20,000 paired bootstrap resamples, whereas prediction bars report descriptive means without confidence intervals.

\paragraph{Training--interface interaction.}
Table~\ref{tab:delay_training_difference} compares clean training, fixed two-step-delay training, and random zero-to-three-step-delay training on Walker. Each training condition contains five checkpoints, each evaluated with ten seeds under both execution interfaces. The adverse average equally weights the six non-clean evaluation conditions. Differences are computed from unrounded returns before reporting rounded values.

\paragraph{Process-separated execution and feedback latency.}
The asynchronous execution stack runs at 10 and 20 Hz on Walker, Cheetah, and Cup, using five paired seeds per condition. Controlled execution shifts include a one-period fixed delay, uniformly sampled integer delays from zero to three periods, 20\% packet loss combined with random delay, and packet reordering implemented by delaying selected packets by two periods with probability 0.60. The feedback-latency analysis uses random command delay and compares immediate, one-period-delayed, and two-period-delayed action feedback.

\paragraph{Metrics and aggregation.}
For the main DreamerV3 results, each task's return is normalized by its mean clean \textsc{CMD} return and then averaged equally across tasks. Appendix Table~\ref{tab:dreamer_main} instead reports the normalized return difference relative to \textsc{CMD}. Recovery in the process-separated stack is computed as
\[
100\frac{R_{\mathrm{corrected}}-R_{\mathrm{baseline}}}
{R_{\mathrm{natural}}-R_{\mathrm{baseline}}},
\]
where natural execution uses the same baseline controller without injected perturbations. Return uncertainty is estimated using complete episodes as the sampling units rather than individual transitions.

\end{document}